\documentclass[11pt,a4paper]{article}
\usepackage[hyperref]{acl2018}
\usepackage{times}
\usepackage{latexsym}

\usepackage{array, caption, floatrow, tabularx, makecell, booktabs}%
\usepackage{hyperref}
\usepackage{color}
\usepackage{colortbl}
\usepackage[colorlinks]{}
\usepackage{subfig}
\usepackage{wrapfig}
\usepackage{graphicx}
\usepackage{url}
\usepackage [autostyle, english = american]{csquotes}
\MakeOuterQuote{"}
\usepackage{fixltx2e}

\aclfinalcopy 

\title{Controlling Personality-Based Stylistic Variation \\ with Neural Natural Language Generators}

\author{Shereen Oraby$^1$, Lena Reed$^1$, Shubhangi Tandon$^1$, \\ 
\bf Sharath T.S.$^1$, \bf Stephanie Lukin$^2$, \bf and Marilyn Walker$^1$ \\
$^1$Natural Language and Dialogue Systems Lab, University of California, Santa Cruz\\
$^2$U.S. Army Research Laboratory, Los Angeles, CA\\
  {\tt \{soraby,lireed,shtandon,sturuvek,mawalker\}@ucsc.edu} \\
  {\tt stephanie.m.lukin.civ@mail.mil} \\}

\date{}

\begin{document}
\maketitle

\begin{abstract}
Natural language generators for task-oriented dialogue must
effectively realize system dialogue actions and their associated
semantics. In many applications, it is also desirable for generators
to control the style of an utterance. To date, work on task-oriented
neural generation has primarily focused on semantic fidelity rather
than achieving stylistic goals, while work on style has been done in
contexts where it is difficult to measure content preservation.  Here
we present three different sequence-to-sequence models and carefully
test how well they disentangle content and style.  We use a
statistical generator, {\sc Personage}, to synthesize a new corpus of
over 88,000 restaurant domain utterances whose style varies according
to models of personality, giving us total control over both the
semantic content and the stylistic variation in the training data. We
then vary the amount of explicit stylistic supervision given to the
three models. We show that our most explicit model can simultaneously
achieve high fidelity to both semantic and stylistic goals: this model
adds a context vector of 36 stylistic parameters as input to the
hidden state of the encoder at each time step, showing the 
benefits of explicit stylistic
supervision, even when the amount of training data is large. 
\end{abstract}

\section{Introduction}

The primary aim of natural language generators (NLGs) for
task-oriented dialogue is to effectively realize system dialogue
actions and their associated content parameters.  This requires
training data that allows the NLG to learn how to map semantic
representations for system dialogue acts to one or more possible
outputs (see Figure \ref{figure:MR}, 
\cite{novikova2016crowd}). Because neural generators often make
semantic errors such as deleting, repeating or hallucinating content,
to date previous work on task-oriented neural generation has primarily
focused on faithfully rendering the meaning of the system's dialogue
act \cite{DusekJ16,Lampouras2016,Mei2015,Wenetal15}.

\begin{figure}[t!bh]
\begin{small}
\begin{tabular}
{@{} p{2.95in} @{}} \toprule
\cellcolor[gray]{0.9} \sc 
inform(name[The Eagle], eatType[coffee shop], food[English],
  priceRange[high], customerRating[average], area[city centre],
  familyFriendly[yes], near[Burger King]) \\ \hline
{\it The three star coffee shop, The Eagle, located near Burger King, gives families a high priced option for English food in the city centre.}   \\ \hline
{\it  Let's see what we can find on The Eagle. Right, The Eagle is a coffee shop with a somewhat average rating. The Eagle is kid friendly, also it's an English restaurant and expensive, also it is near Burger King in the city centre, you see?}   \\ \hline
\end{tabular}
\vspace{-.1in}
 \caption{Dialogue Act Meaning Representation (MR) with content parameters
and outputs \label{figure:MR}}
\end{small}
\end{figure}

\begin{table*}[t!]
\begin{footnotesize}
\begin{tabular}
{@{} p{2.5cm}|p{13cm} @{}} \toprule
\textbf{Personalities} & \textbf{Realization}  \\ \midrule 
{\cellcolor[gray]{0.9}\sc Meaning}    & {\cellcolor[gray]{0.9} \sc name[Fitzbillies], eatType[pub], food[Italian], priceRange[moderate], } \\ 
{\cellcolor[gray]{0.9}\sc Representation (MR)}    & {\sc \cellcolor[gray]{0.9} customer rating[decent], area[riverside], familyFriendly[no],near['The Sorrento']} \\  \midrule 
{\sc No-Agg/\newline No-Prag} & Fitzbillies is a pub. Fitzbillies has a decent rating. Fitzbillies is moderately priced. Fitzbillies is in riverside. Fitzbillies is an Italian restaurant. Fitzbillies is not family friendly. Fitzbillies is near The Sorrento.\\ \hline
{\sc Agreeable}  &  Let's see what we can find on Fitzbillies. I see, well it is a pub with a decent rating, also it is an Italian restaurant in riverside and moderately priced near The Sorrento, also it isn't family friendly, you see? \\ \hline
{\sc Disagreeable} & I mean, everybody knows that moderately priced Fitzbillies is in riverside with a decent rating. It's near The Sorrento. It isn't family friendly. It is an Italian place. It is a pub. \\ \hline
{\sc Conscientious}  &  Let's see what we can find on Fitzbillies. I see, well it is a pub with a decent rating, it isn't kid friendly and it's moderately priced near The Sorrento and an Italian restaurant in riverside.  \\ \hline
{\sc Unconscientious}    &  Oh god yeah, I don't know. Fitzbillies is a pub with a decent rating, also it is moderately priced near The Sorrento and an Italian place in riverside and it isn't kid friendly. \\ \hline
{\sc Extravert}  & Basically, Fitzbillies is an Italian place near The Sorrento and actually moderately priced in riverside, it has a decent rating, it isn't kid friendly and it's a pub, you know.    \\ \hline
\end{tabular}
\vspace{-.1in}
\caption{Sample neural model output realizations for the same MR for {\sc Personage}'s personalities}
\label{table:personality_sentences}
\end{footnotesize}
\end{table*}







However, in many applications it is also desirable for generators to
control the style of an utterance independently of its content.  For
example, in Figure~\ref{figure:MR}, the first output uses more formal
language and complex syntactic structures, as one might see in
written language, while the other uses simpler syntax and pragmatic
markers characteristic of oral language \cite{Biber91}.  In this paper, we
create several different sequence-to-sequence models and compare
how well they can disentangle content and style.  Controlling the
style of the output requires disentangling the content from the style,
but previous neural models aimed at achieving stylistic goals have not
focused on task-oriented dialogue where specific semantic attributes
and values must be communicated (as per the MR in
Figure~\ref{figure:MR}), and where semantic fidelity can be precisely
measured.\footnote{We
  leave a detailed review of related work to
  Section~\ref{related-sec}.}

One of the main challenges is the lack of parallel corpora realizing
the same content with different styles.  Thus we 
 create a large, novel parallel corpus with specific style
parameters and specific semantics, by using an existing statistical
generator, {\sc Personage} \cite{MairesseWalker10}, to synthesize over 88,000 utterances in
the restaurant domain that vary in style according to psycholinguistic
models of personality.\footnote{Our stylistic
  variation for NLG corpus is available at: {\tt
    nlds.soe.ucsc.edu/stylistic-variation-nlg}} {\sc Personage} can
generate a very large number of stylistic variations for any given
dialogue act, thus yielding, to our knowledge, the largest
style-varied NLG training corpus in existence.  The strength of this
new corpus is that: (1) we can use the {\sc Personage} generator to
generate as much training data as we want; (2) it allows us to
systematically vary a specific set of stylistic parameters and the
network architectures; (3) it allows us to systematically test the
ability of different models to generate outputs that faithfully
realize both the style and content of the training data.\footnote{
Section~\ref{results-sec}  quantifies the 
naturalness of {\sc Personage} outputs.}

We develop novel neural models that vary the amount of explicit
stylistic supervision given to the network, and we explore, for the
first time, explicit control of multiple interacting stylistic
parameters. We show that the no-supervision ({\sc no-sup}) model, a
baseline sequence-to-sequence model
\cite{sutskever2014sequence,DusekJ16}, produces semantically correct
outputs, but eliminates much of the stylistic variation that it saw in
the training data. {\sc Model\_Token} provides minimal
supervision by allocating a latent variable in the encoding as a label
for each style, similar to the use of language labels in machine
translation \cite{johnson2016google}.  This model learns to generate
coherent and stylistically varied output without explicit exposure to
language rules, but makes more semantic errors. {\sc Model\_Context} adds 
another layer to provide an additional encoding of
individual stylistic parameters to the network. We show that it
performs best on both measures of semantic fidelity and stylistic
variation. The results suggest that neural architectures can benefit
from explicit stylistic supervision, even with a large training set.

\section{Corpus Creation} 
\label{Data}

We aim to systematically create a corpus that can be used to test how
different neural architectures affect the ability of the trained model
to disentangle style from content, and faithfully produce semantically
correct utterances that vary style. We use {\sc Personage}, an
existing statistical generator: due to space, we briefly explain how
it works, referring the interested reader to \newcite{MairesseWalker10,MairesseWalker11} for details.


{\sc Personage} requires as input: (1) a meaning representation (MR)
of a dialogue act and its content parameters, and (2) a parameter file
that tells it how frequently to use each of its stylistic
parameters. Sample model outputs are shown in the second row of Figure~\ref{figure:MR} and in 
Table~\ref{table:personality_sentences}, illustrating
some stylistic variations {\sc personage} produces. 

To generate our novel corpus, we utilize the MRs from the E2E Generation
Challenge.\footnote{{\url{http://www.macs.hw.ac.uk/InteractionLab/E2E/}}}
The MR in Figure~\ref{figure:MR} illustrates {\bf all 8}
available attributes. We added a dictionary entry for each attribute
to {\sc personage} so that it can express that
attribute.\footnote{{\sc personage} supports a one-to-many mapping
  from attributes to elementary syntactic structures for expressing
  that attribute, but here we use only one dictionary
  entry. {\sc personage} also allows
  for discourse relations such as justification or contrast to hold
  between content items, but the E2E MRs do not include such
  relations.}  These dictionary entries are syntactic representations
for very simple sentences: the {\sc no-agg/no-prag} row of
Table~\ref{table:personality_sentences} shows a sample realization of
each attribute in its own sentence based on its dictionary entry.

\begin{table}[!htb]
\begin{footnotesize}
\begin{tabular}
{@{} p{.45in}|p{0.2in}|p{0.2in}|p{0.2in}|p{0.2in}|p{0.2in}|p{0.2in}@{}}
\hline
& \multicolumn{6}{c}{ {\bf Number of Attributes in MR}}  \\       \hline        
{\bf  Dataset} & {\bf  3} & {\bf  4} & {\bf 5} & {\bf  6} & {\bf 7} & {\bf  8} \\ \hline 
{\sc  Train} & 0.13 & 0.30 & 0.29 & 0.22 & 0.06 & 0.01 \\ \hline
{\sc  Test} & 0.02 & 0.04 & 0.06 & 0.15 & 0.35 & 0.37 \\
\hline
\end{tabular}
\end{footnotesize}
\
\vspace{-.1in}
\centering \caption{\label{table:mr-dst} Percentage of the MRs in training and test
in terms of number of attributes in the MR}
\end{table}

We took advantage of the setup of the E2E Generation Challenge and
used their MRs, exactly duplicating their split between training, dev
and test MRs, because they ensured that the dev and test MRs had not
been seen in training.  The frequencies of longer utterances (more
attribute MRs) vary across train and test, with actual distributions
in Table \ref{table:mr-dst}, showing how the test set was designed to
be challenging, while the test set in \newcite{Wenetal15} averages
less than 2 attributes per MR \cite{Nayaketal17}.  We combine their
dev and training MRs resulting in 3784 unique MRs in the training set,
and generate 17,771 reference utterances per personality for a
training set size of 88,855 utterances.  The test set consists of 278
unique MRs and we generate 5 references per personality for a test
size of 1,390 utterances.



\begin{table}[htb!]
\begin{footnotesize}
\begin{tabular}
{@{} p{1.2in}|p{1.65in} @{}}
\hline
{\bf Attribute} & {\bf Example} \\ \hline\hline
\multicolumn{2}{l}{ \cellcolor[gray]{0.9} {\sc Aggregation Operations}}     \\               
{\sc Period} &  {\it X serves Y. It is in  Z.}  \\
{\sc "With" cue} &  {\it X is in Y, with Z.}  \\
{\sc Conjunction} & {\it X is Y and it is Z. \& X is Y, it is Z.} \\
{\sc All Merge} & {\it X is Y, W and Z \& X is Y in Z}  \\
{\sc "Also" cue} &  {\it X has Y, also it has Z.} \\
\multicolumn{2}{l}{ \cellcolor[gray]{0.9} {\sc Pragmatic Markers}}     \\    
{\sc ack\_definitive} & \it right, ok \\ 
{\sc ack\_justification} & \it I see, well \\ 
{\sc ack\_yeah} & \it yeah\\ 
{\sc confirmation} & \it let's see what we can find on X, let's see ....., did you say X?  \\ 
{\sc initial rejection} & \it mmm, I'm not sure, I don't know.  \\ 
{\sc competence mit.} & \it come on, obviously, everybody knows that \\ 
{\sc filled pause stative} & \it err, I mean, mmhm \\ 
{\sc down\_kind\_of} & \it kind of \\ 
{\sc down\_like} & \it like \\ 
{\sc down\_around} & \it around \\  
{\sc exclaim} & \it ! \\
{\sc indicate surprise} & \it oh \\
{\sc general softener} & \it sort of, somewhat, quite, rather \\ 
{\sc down\_subord} & \it I think that, I guess \\ 
{\sc emphasizer} & \it really, basically, actually, just \\ 
{\sc emph\_you\_know} & \it you know \\ 
{\sc expletives} & \it oh god, damn, oh gosh, darn  \\ 
{\sc in group marker} & \it pal, mate, buddy, friend \\ 
{\sc tag question} & \it alright?, you see? ok? \\ 
\hline
\end{tabular}
\end{footnotesize}
\vspace{-.1in}
\centering \caption{\label{table:agg-prag} Aggregation and Pragmatic Operations}
\end{table}

The experiments are based on two types of parameters provided with
{\sc personage}: aggregation parameters and pragmatic 
parameters.\footnote{We disable parameters related to content
  selection, syntactic template selection and lexical choice.}  The
{\sc no-agg/no-prag}  row of Table~\ref{table:personality_sentences} shows
what {\sc Personage} would output if it did not use any of its
stylistic parameters.  
The top half of
Table~\ref{table:agg-prag} illustrates the aggregation parameters:
these parameters control how the NLG combines attributes into
sentences, e.g., whether it tries to create complex sentences by
combining attributes into phrases and what types of combination
operations it uses.  The pragmatic operators are shown in the second
half of Table~\ref{table:agg-prag}. Each parameter value can be set to {\tt high},
{\tt low}, or {\tt don't care}. 

\begin{figure}[t!]
    \centering
    \caption{Frequency of the Top 2 most frequent Aggregation and Pragmatic Markers in Train}
        \subfloat[Aggregation Operations]{\includegraphics[width=\columnwidth]{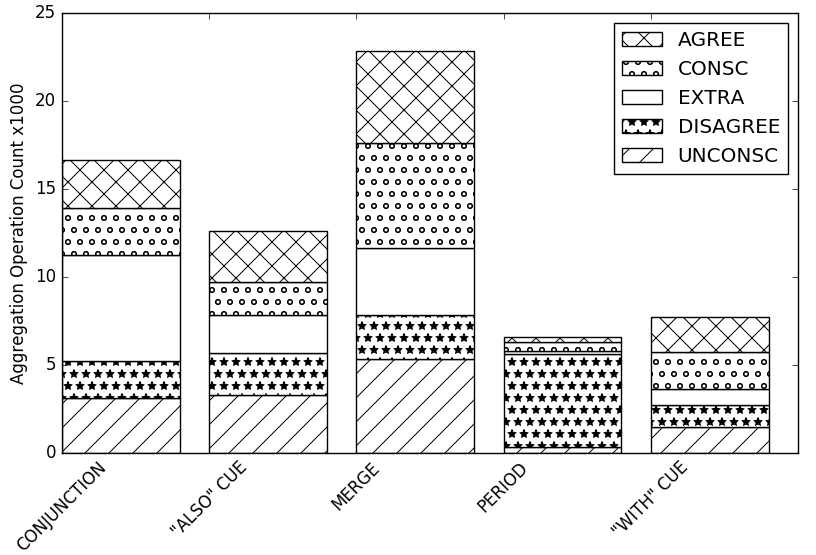}}
\vspace{-.15in}        
        \subfloat[Pragmatic Markers]{\includegraphics[width=\columnwidth]{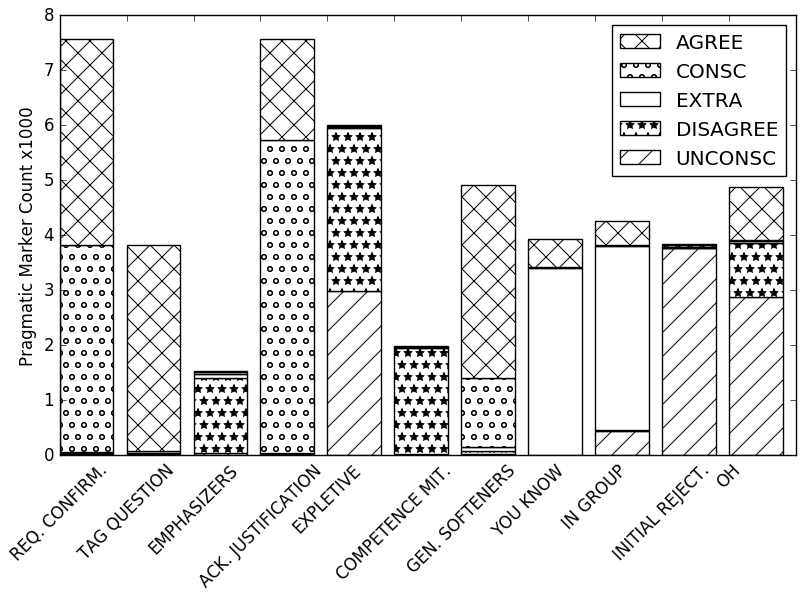}}
    \label{fig:agg_prag_plot}
\end{figure}

To use {\sc Personage} to create training data mapping the same MR to
multiple personality-based variants, we set {\bf values} for
{\bf all} of the parameters in Table~\ref{table:agg-prag} using the
stylistic models defined by \newcite{MairesseWalker10} for the
following Big Five personality traits: agreeable, disagreeable,
conscientiousness, unconscientiousness, and
extravert. Figure~\ref{fig:agg_prag_plot} shows that each personality
produces data that represents a stylistically distinct
distribution. These models are probabilistic and specified values are
automatically broadened within a range, thus each model can produce
10's of variations for each MR. Note that while each personality type
distribution can be characterized by a single stylistic label (the
personality), Figure~\ref{fig:agg_prag_plot} illustrates that each
distribution is characterized by multiple interacting
stylistic parameters. 

Each parameter modifies the linguistic structure in order to create
distributionally different subcorpora.  To see the effect of each
personality using a different set of aggregation operators,
cross-reference the aggregation operations in
Table~\ref{table:agg-prag} with an examination of the outputs in
Table~\ref{table:personality_sentences}.  The simplest choice for
aggregation does not combine attributes at all: this is represented by
the {\sc period} operator, which, if used persistently, results in an
output with each content item in its own sentence as in the {\sc
  no-agg/no-prag} row, or the content being realized over multiple
sentences as in the {\sc disagreeable} row (5 sentences).  However, if
the other aggregation operations have a high value, {\sc personage}
prefers to combine simple sentences into complex ones whenever it can,
e.g., the {\sc extravert} personality example in
Table~\ref{table:personality_sentences} combines all the attributes
into a single sentence by repeated use of the {\sc all merge} and {\sc
  conjunction} operations. The {\sc conscientious} row in
Table~\ref{table:personality_sentences} illustrates the use of the
{\sc with-cue} aggregation operation, e.g., {\it with a decent rating}.
Both the {\sc agreeable} and {\sc conscientious} rows in
Table~\ref{table:personality_sentences} provide examples of the {\sc
  also-cue} aggregation operation.  In {\sc personage}, the
aggregation operations are defined as syntactic operations on the
dictionary entry's syntactic tree. Thus to mimic these operations
correctly, the neural model must derive latent representations that
function as though they also operate on syntactic trees.

The pragmatic operators in the second half of
Table~\ref{table:agg-prag} are intended to achieve particular
pragmatic effects in the generated outputs: for example the use of a
hedge such as {\it sort of} softens a claim and affects perceptions of
friendliness and politeness \cite{BrownLevinson87}, while the
exaggeration associated with emphasizers like {\it actually,
  basically, really} influences perceptions of extraversion and
enthusiasm \cite{OberlanderGill04b,DewaeleFurnham99}. In {\sc
  personage}, the pragmatic parameters are attached to the syntactic
tree at {\it insertion points} defined by syntactic constraints,
e.g., {\sc emphasizers} are adverbs that can occur sentence initially
or before a scalar adjective.  Each personality model uses a
variety of pragmatic parameters. Figure~\ref{fig:agg_prag_plot} shows how
these markers distribute differently across personality models, with
examples in Table
\ref{table:personality_sentences}.

\section{Model Architectures}
\label{Models}

Our neural generation models build on the open-source
sequence-to-sequence (seq2seq) TGen system
\cite{DusekJ16a}\footnote{\url{https://github.com/UFAL-DSG/tgen}},
implemented in Tensorflow \cite{Abadi16}.  The system is based on 
seq2seq generation with attention
\cite{DBLP:journals/corr/BahdanauCB14, sutskever2014sequence}, and
uses a sequence of LSTMs \cite{hochreiter1997long} for the encoder and
decoder, combined with beam-search and reranking for
output tuning.

The input to TGen are dialogue acts for each system action (such as {\it
  inform}) and a set of attribute slots (such as {\it rating}) and
their values (such as {\it high} for attribute {\it rating}).  The
system integrates sentence planning and surface realization into a
single step to produce natural language outputs.  To preprocess the
corpus of MR/utterance pairs, attributes that take on proper-noun
values are delexicalized during training i.e., {\it name} and {\it
  near}.  During the generation phase on the test set, a
post-processing step re-lexicalizes the outputs. The MRs (and
resultant embeddings) are sorted internally by dialogue act tag and
attribute name.

\begin{figure}[t!h]
\centering
\includegraphics[width=0.95\linewidth, keepaspectratio]{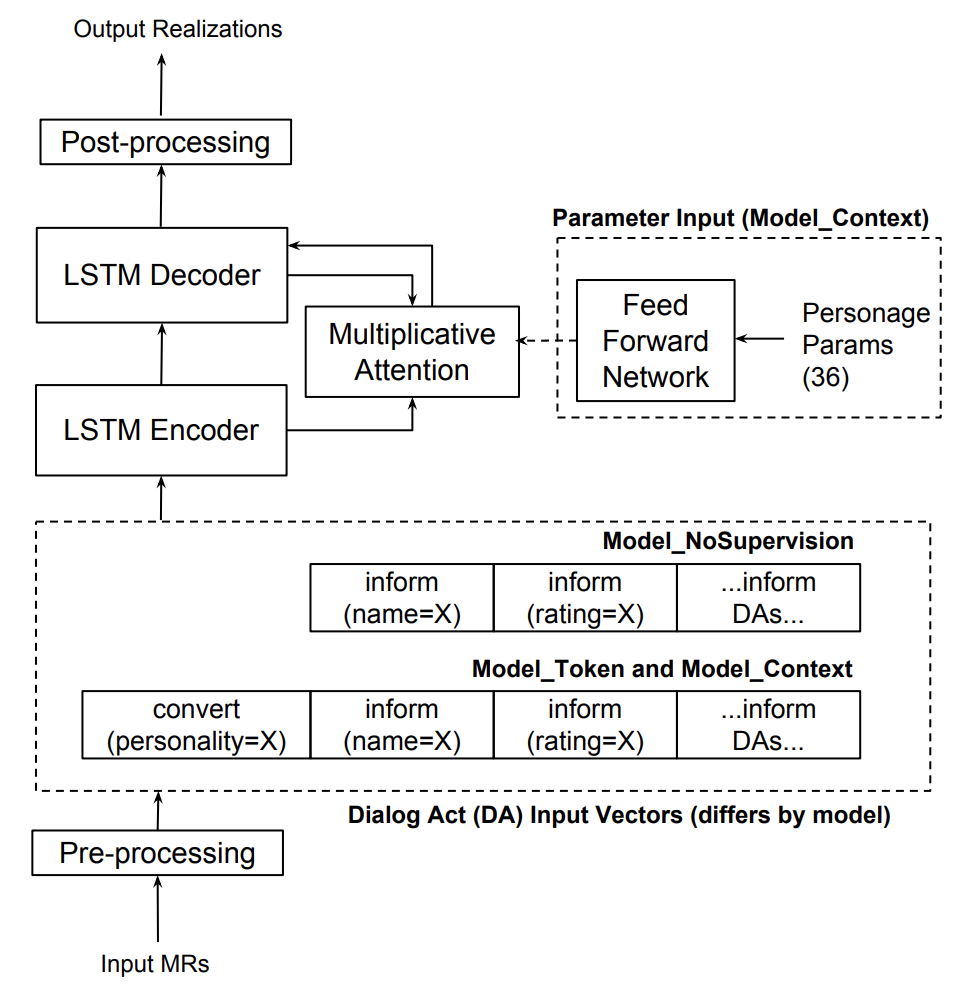}
\vspace{-.15in}
\caption{\label{fig:model1}Neural Network Model Architecture}
\end{figure}

The models are designed to systematically test the effects of increasing
the level of supervision, with novel architectural
additions to accommodate these changes. We use the
default parameter settings from TGen \cite{DusekJ16a} with batch size
20 and beam size 10, and use 2,000 training instances for
parameter tuning to set the number of training epochs and learning rate. Figure
\ref{fig:model1} summarizes the architectures.

\vspace{0.1cm}
{ \noindent{\bf {\sc Model\_NoSupervision}.}
The simplest model follows the baseline TGen architecture \cite{DusekJ16},
with training using all 88K utterances in a single pool
for up to 14 epochs based on loss monitoring for the decoder and reranker.

\vspace{0.1cm}
\noindent{\bf {\sc Model\_Token}.}  The second model adds a token of
additional supervision by introducing a new dialogue act, {\it
  convert}, to encode personality, inspired by the use of a language
token for machine translation \cite{johnson2016google}.  Unlike other
work that uses a single token to control generator output
\cite{fancontrollable,hu2017toward}, the personality token encodes a
constellation of different parameters that define the style of the
matching reference.  Uniquely here, the model attempts to {\it
  simultaneously} control multiple style variables that may interact
in different ways. Again, we monitor loss on the validation set and
training continues for up to 14 epochs for the decoder and reranker.

\vspace{0.1cm}
\noindent{\bf {\sc Model\_Context}.}  The most complex model
introduces a context vector, as shown at the top right of Figure
\ref{fig:model1}.
The vector  explicitly encodes a set of 36 style parameters
from Table \ref{table:agg-prag}. The
parameters for each reference text are encoded as a boolean vector,
and a feed-forward network is added as a context encoder, taking the
vector as input to the hidden state of the encoder and making the
parameters available at every time step to a multiplicative attention
unit. The activations of the fully
connected nodes are represented as an additional time step of the
encoder of the seq2seq architecture \cite{sutskever2014sequence}. The
attention \cite{DBLP:journals/corr/BahdanauCB14} is computed over all
of the encoder states and the hidden state of the fully connected
network.  Again, we set the learning rate, alpha decay, and maximum training epochs
(up to 20) based on loss monitoring on the validation set.

\section{Quantitative Results}
\label{results-sec}

Here, we present results on controlling stylistic variation while maintaining semantic fidelity.

\subsection{Evaluating Semantic Quality}
\label{eval:semantic-quality}
It is widely agreed that new evaluation metrics are needed for NLG
\cite{Langkilde2002,BelzReiter06,BRW00,novikovanewnlg}. We first present automated
metrics used in NLG to measure how well model outputs compare to
{\sc Personage} input, then introduce novel metrics designed to
fill the gap left by current evaluation metrics.

\noindent {\bf Automatic Metrics.} The automatic evaluation uses the
E2E generation challenge script.\footnote{\url{https://github.com/tuetschek/e2e-metrics}} Table~\ref{table:auto_metrics}
summarizes the results for BLEU (n-gram precision), NIST (weighted n-gram precision),
METEOR (n-grams with synonym recall), and ROUGE (n-gram
recall). Although the differences in metrics are small, {\sc
  Model\_context}
shows a slight improvement across all of the metrics.


\begin{table}[!htb]
\begin{footnotesize}
\begin{tabular}
{@{} p{0.45in}|p{0.45in}|p{0.45in}|p{0.45in}|p{0.45in} @{}}
\hline
{\bf \scriptsize Model} & {\bf \scriptsize BLEU} & {\bf \scriptsize  NIST} & {\bf \scriptsize METEOR} & {\bf \scriptsize ROUGE\_L} \\ \hline     
{\sc \scriptsize NoSup } & 0.2774  & 4.2859 & 0.3488 & 0.4567  \\
{\sc \scriptsize Token } & 0.3464  & 4.9285 & 0.3648 & 0.5016  \\
{\sc \scriptsize Context } & \bf 0.3766  & \bf 5.3437 & \bf 0.3964 & \bf 0.5255  \\
\hline
\end{tabular}
\end{footnotesize}
\vspace{-.1in}
\centering \caption{\label{table:auto_metrics} Automated Metric Evaluation}
\end{table}

                                   
\noindent {\bf Deletions, Repetitions, and Substitutions.}  Automated
evaluation metrics are not informative about the quality of the
outputs, and penalize models for introducing stylistic variation. We
thus develop new scripts to automatically evaluate the types 
common types of neural generation errors: {\it deletions} (failing to realize a value), {\it
  repeats} (repeating a value), and {\it
  substitutions} (mentioning an attribute with an incorrect value).


Table \ref{table:deletes-repeats} shows ratios for the number of
deletions, repeats, and substitutions for each model for the test set
of 1,390 total realizations (278 unique MRs, each realized once for
each of the 5 personalities). The error counts are split by
personality, and normalized by the number of unique MRs (278). 
Smaller ratios are preferable, indicating fewer errors. Note 
that because {\sc Model\_NoSup} does not encode a
personality parameter, the error values are the same
across each personality (averages across the full test set).
 
The table shows that  {\sc Model\_NoSup}
makes very few semantic errors (we show later that this is at the cost of limited
stylistic variation). Across all error types, {\sc Model\_Context}
makes significantly fewer errors than {\sc Model\_Token},
suggesting that its additional explicit parameters
help avoid semantic errors. 
The last row quantifies whether some personalities
are harder to model: it  shows that across all models,
{\sc disagreeable} and {\sc extravert} have the most errors,
while  {\sc conscientious} has the fewest.

\begin{table}[!htb]
\begin{footnotesize}
\begin{tabular}
{@{} p{0.35in}|p{0.3in}|p{0.3in}|p{0.35in}|p{0.35in}|p{0.38in} @{}}
\hline
{\bf \scriptsize Model} & {\bf \scriptsize AGREE} & {\bf \scriptsize  CONSC} & {\bf \scriptsize DISAG} & {\bf \scriptsize EXTRA} & {\bf \scriptsize UNCON} \\ \hline\hline  
\multicolumn{6}{l}{ \cellcolor[gray]{0.9} {\sc Deletions}}     \\ 
{\sc \scriptsize NoSup } & \bf 0.01  & \bf 0.01 & \bf 0.01 &  \bf 0.01 &\bf  0.01 \\                            
{\sc \scriptsize Token } & 0.27  & 0.22 & 0.87 & 0.74  & 0.31 \\
{\sc \scriptsize Context } & 0.08 & \bf 0.01 & 0.14 & 0.08 & \bf 0.01 \\\hline\hline  
\multicolumn{6}{l}{ \cellcolor[gray]{0.9} {\sc Repetitions}}     \\                             
{\sc \scriptsize NoSup } & \bf 0.00  & \bf 0.00 & \bf 0.00 & \bf 0.00  & \bf 0.00 \\                            
{\sc \scriptsize Token } & 0.29  & 0.12 & 0.81 & 0.46  & 0.28 \\
{\sc \scriptsize Context } & 0.02 & \bf 0.00 & 0.14 & \bf 0.00 & \bf 0.00 \\\hline\hline
\multicolumn{6}{l}{ \cellcolor[gray]{0.9} {\sc Substitutions}}     \\                     
{\sc \scriptsize NoSup } & 0.10  & 0.10 & 0.10 & 0.10  & 0.10 \\                            
{\sc \scriptsize Token } &  0.34 & 0.41 & 0.22 & 0.35  & 0.29 \\
{\sc \scriptsize Context } & \bf 0.03  & \bf 0.03 & \bf 0.00 & \bf 0.00 & \bf 0.03 \\\hline \hline
{\bf All} & 0.68 & 0.35 & 1.96 & 1.29 & 0.61  \\ \hline
\end{tabular}
\end{footnotesize}
\vspace{-.1in}
\centering \caption{\label{table:deletes-repeats} Ratio of Model Errors by Personality}
\end{table}

\subsection{Evaluating Stylistic Variation}

Here we characterize the fidelity of stylistic
variation across different model outputs.

\noindent {\bf Entropy.} Shannon text entropy quantifies the amount of
variation in the output produced by each model.  We calculate entropy as $-\sum_{x\in S} \frac{freq}{total} *
log_2(\frac{freq}{total})$, where $S$ is the set of unique words in
all outputs generated by the model, $freq$ is the frequency of
a term, and $total$ counts the number of terms in all references.
Table~\ref{table:entropy} shows that the training
data has the highest entropy, but {\sc Model\_Context} performs the best at
preserving the variation seen in the training data. 
Row {\sc NoSup} shows that {\sc Model\_NoSup} makes the
fewest semantic errors, but produces the least varied output. {\sc
  Model\_Context}, informed by the explicit stylistic context encoding, makes
comparably few semantic errors, while producing
stylistically varied output with high entropy.

\begin{table}[!htb]
\begin{footnotesize}
\begin{tabular}
{@{} p{0.9in}|p{0.45in}|p{0.45in}|p{0.45in} @{}}
\hline
{\bf \scriptsize Model} & {\bf \scriptsize 1-grams} & {\bf \scriptsize  1-2grams} & {\bf \scriptsize 1-3grams} \\ \hline     
{\sc \scriptsize {\sc Personage}Train } & 5.97  & 7.95 & 9.34 \\ \hline\hline
{\sc \scriptsize NoSup } & 5.38  & 6.90 & 7.87  \\
{\sc \scriptsize Token } & 5.67  & 7.35 & 8.47  \\
{\sc \scriptsize Context } & \bf 5.70  & \bf 7.42 & \bf 8.58 \\
\hline
\end{tabular}
\end{footnotesize}
\vspace{-.1in}
\centering \caption{\label{table:entropy} Shannon Text Entropy}
\end{table}

\vspace{0.1cm}
\noindent {\bf Pragmatic Marker Usage.}  To measure whether the
trained models faithfully reproduce the pragmatic markers for each
personality, we count each pragmatic marker in Table
\ref{table:agg-prag} in the output, average the counts and compute the
Pearson correlation between the {\sc personage} references and the
outputs for each model and personality. See Table
\ref{table:prag-results} (all correlations significant with $p \le
0.001$).

\begin{table}[!htb]
\begin{footnotesize}
\begin{tabular}
{@{} p{0.35in}|p{0.3in}|p{0.3in}|p{0.35in}|p{0.35in}|p{0.38in} @{}}
\hline
{\bf \scriptsize Model} & {\bf \scriptsize AGREE} & {\bf \scriptsize  CONSC} & {\bf \scriptsize DISAG} & {\bf \scriptsize EXTRA} & {\bf \scriptsize UNCON} \\ \hline                         
{\sc \scriptsize NoSup } & 0.05  & 0.59 & -0.07 & -0.06 & -0.11  \\
{\sc \scriptsize Token } & \bf 0.35  & 0.66 & 0.31 & 0.57 & 0.53  \\
{\sc  \scriptsize Context } & 0.28 &  \bf 0.67 & \bf 0.40 & \bf 0.76 &\bf  0.63  \\
\hline
\end{tabular}
\end{footnotesize}
\vspace{-.1in}
\centering \caption{\label{table:prag-results} Correlations between {\sc Personage} and models for pragmatic markers in Table \ref{table:agg-prag}}
\end{table}

Table \ref{table:prag-results} shows that {\sc Model\_Context} has the
highest correlation with the training data, for all personalities
(except {\sc agreeable}, with significant margins, and {\sc
  conscientious}, which is the easiest personality to model, with a margin of 0.01).  
  While {\sc Model\_NoSup} shows
positive correlation with {\sc agreeable} and {\sc conscientious}, it
shows {\it negative} correlation with the {\sc Personage} 
inputs for {\sc disagreeable}, {\sc extravert}, and {\sc
  unconscientious}. The pragmatic marker distributions for {\sc
  Personage} train in Figure \ref{fig:agg_prag_plot} indicates that
the {\sc conscientious} personality most frequently uses {\it
  acknowledgement-justify} (i.e., {\it "well"}, {\it "i see"}), and
{\it request confirmation} (i.e., {\it "did you say X?"}), which are
less complex to introduce into a realization since they often lie at
the beginning or end of a sentence, allowing the simple {\sc
  Model\_NoSup} to learn them.\footnote{We verified
that there is not a high
  correlation between every set of pragmatic markers:
  different personalities do not correlate, e.g., -0.078 for {\sc
    Personage} {\sc disagreeable} and {\sc Model\_Token} {\sc
    agreeable}.}

\noindent {\bf Aggregation.}\label{eval:aggreg} To measure the ability
of each model to aggregate, we average the
counts of each aggregation operation for each model and
personality and compute the Pearson correlation between the output and the {\sc
  personage} training data. 
\begin{table}[!htb]
\begin{footnotesize}
\begin{tabular}
{@{} p{0.35in}|p{0.3in}|p{0.3in}|p{0.35in}|p{0.35in}|p{0.38in} @{}}
\hline
{\bf \scriptsize Model} & {\bf \scriptsize AGREE} & {\bf \scriptsize  CONSC} & {\bf \scriptsize DISAG} & {\bf \scriptsize EXTRA} & {\bf \scriptsize UNCON} \\ \hline                           
{\sc \scriptsize NoSup } & 0.78 & 0.80 & 0.13 & 0.42 & 0.69 \\
{\sc \scriptsize Token } & 0.74 & 0.74 & \bf 0.57 & 0.56 & 0.60 \\
{\sc \scriptsize Context } & \bf 0.83 & \bf 0.83 & 0.55 & \bf 0.66 & \bf 0.70 \\
\hline
\end{tabular}
\end{footnotesize}
\vspace{-.1in}
\centering \caption{\label{table:aggreg-results} Correlations between {\sc Personage} and models for aggregation operations in Table \ref{table:agg-prag}}
\end{table}

The correlations in
Table \ref{table:aggreg-results} (all significant
with $p \le 0.001$) show that {\sc Model\_Context} has a higher
correlation with {\sc Personage} than the two simpler models
(except for {\sc disagreeable}, where {\sc Model\_Token} is higher
by 0.02). Here, {\sc Model\_NoSup} actually {\it
  frequently} outperforms the more informed {\sc Model\_Token}. 
Note that {\it all personalities use aggregation}, even thought
{\it {\bf not} all personalities use pragmatic markers}, and so even
without a special {\it personality} token, {\sc Model\_NoSup}
is able to faithfully reproduce aggregation operations. In fact, since the correlations are
frequently higher than those for {\sc Model\_Token}, we hypothesize
that is able to more accurately focus on aggregation (common to all
personalities) than stylistic differences, which {\sc Model\_Token}
is able to produce.

\section{Qualitative Analysis}
\label{quality}

Here, we present two evaluations aimed at qualitative analysis of our outputs.

\vspace{.1cm}
\noindent {\bf Crowdsourcing Personality Judgements. }\label{eval:mturk}
Based on our quantitative results, we select {\sc Model\_Context} as the best-performing model
and conduct an evaluation to test if humans can distinguish the personalities exhibited.
We randomly select a set of 10 unique MRs from the {\sc Personage}
training data along with their corresponding reference texts for each
personality (50 items in total), and 30 unique MRs {\sc
  Model\_Context} outputs (150 items in total).\footnote{Note that we
  use fewer {\sc Personage} references simply to validate that our
  personalities are distinguishable in training, but will
  more rigorously evaluate our model in future work.}  We construct a HIT on 
Mechanical Turk, presenting a
single output (either {\sc Personage} or {\sc Model\_Context}), and ask
5 Turkers to label the output using the Ten Item Personality Inventory (TIPI)
\cite{gosling2003tipi}. The TIPI is a ten-item measure of the Big Five
personality dimensions, consisting of two items for each of the five
dimensions, one that {\it matches} the dimension, and one that is the
{\it reverse} of it, and a scale that ranges from 1 (disagree
strongly) to 7 (agree strongly). To qualify Turkers for the task, we
ask that they first complete a TIPI on themselves, to help ensure that they
understand it.

\begin{table}[!htb]
\begin{footnotesize}
\begin{tabular}
{@{} p{0.35in}|p{0.25in}|p{0.25in}|p{0.25in}|p{0.25in}|p{0.25in}|p{0.25in}@{}}
\hline
& \multicolumn{3}{c|}{{\sc Personage}}  &  \multicolumn{3}{c}{{\sc Model\_Context}} \\ \hline                           
{\bf \scriptsize Person.} & {\bf \scriptsize  Ratio \newline Correct} & {\bf \scriptsize Avg. \newline Rating} & {\bf \scriptsize Nat. \newline Rating} & {\bf \scriptsize  Ratio \newline Correct} & {\bf \scriptsize Avg. \newline Rating} & {\bf \scriptsize Nat. \newline Rating} \\ \hline                           
{\sc \scriptsize AGREE } & 0.60 & 4.04 & 5.22 & 0.50 & 4.04 & 4.69\\
{\sc \scriptsize DISAGR } & 0.80 & 4.76 & 4.24 & 0.63 & 4.03 & 4.39 \\
{\sc \scriptsize CONSC } & 1.00 & 5.08 & 5.60 & 0.97 & 5.19 & 5.18 \\
{\sc \scriptsize UNCON } & 0.70 & 4.34 & 4.36 & 0.17 & 3.31 & 4.58 \\
{\sc \scriptsize EXTRA } & 0.90 & 5.34 & 5.22 & 0.80 & 4.76 & 4.61 \\
\hline
\end{tabular}
\end{footnotesize}
\vspace{-.1in}
\centering \caption{\label{table:mturk} Percentage of Correct Items and Average Ratings and Naturalness Scores for Each Personality ({\sc Personage} vs. {\sc Model\_Context})}
\end{table}

Table \ref{table:mturk} presents results as aggregated counts for the
number of times at least 3 out of the 5 Turkers rated the {\it
  matching} item for that personality higher than the {\it reverse}
item ({Ratio Correct}), the average rating the correct item
received (range between 1-7), and an average "naturalness" score for
the output (also rated 1-7). From the table, we can see that for {\sc
  Personage} training data, all of the personalities have a correct ratio that is higher than 0.5. The {\sc Model\_Context}
outputs exhibit the same trend except for {\sc unconscientious} and
{\sc agreeable}, where the correct ratio is only 0.17 and 0.50,
respectively (they also have the lowest correct ratio for the original {\sc Personage} data).

Table \ref{table:mturk} also presents results for naturalness for both
the reference and generated utterances, showing that both achieve decent
scores for naturalness (on a scale of 1-7). While human
utterances would probably be judged more natural, it is not at all
clear that similar experiments could be done with human generated
utterances, where it is difficult to enforce the same amount of
experimental control.

\vspace{0.1cm}
\noindent{\bf Generalizing to Multiple Personalities.}  A final
experiment explores whether the models learn additional stylistic
generalizations not  seen in training. We train a version of {\sc Model\_Token}, as before on instances
with single personalities, but such that it can be used to generate
output with a combination of {\it two} personalities. The experiment
uses the original training data for {\sc Model\_Token}, but uses an
expanded test set where the MR includes {\bf two} personality {\sc
  convert} tags. We pair each personality with all personalities
except its exact opposite.


\begin{table}[t!]
\begin{scriptsize}
\begin{tabular}
{@{} p{0.1in}|p{0.3in}|p{0.25in}|p{0.25in}|p{1.35in}@{}}
\toprule
 & {\bf \scriptsize  Persona} & {\bf \scriptsize Period Agg.} & {\bf \scriptsize Explet Prag.} & {\bf \scriptsize  Example} \\ \hline          
\rowcolor [gray]{0.9}1 & DISAG & 5.71 & 2.26 &  Browns Cambridge is damn moderately priced, also it's in city centre. It is a pub. It is an italian place. It is near Adriatic. It is damn family friendly. \\ \hline\hline              
2 & CONSC            & 0.60  & 0.02 & Let's see what we can find on Browns Cambridge. I see, well it is a pub, also it is moderately priced, an italian restaurant near Adriatic and family friendly in city centre. \\\hline
\rowcolor [gray]{0.9}3 & DISAG+\newline CONSC   & 3.81 & 0.84 & Browns Cambridge is an italian place and moderately priced. It is near Adriatic. It is kid friendly. It is a pub. It is in riverside. \\\bottomrule
\end{tabular}
\end{scriptsize}
\vspace{-.1in}
\centering \caption{\label{table:multi-voice-examples} {Multiple-Personality Generation Output based on {\sc Disagreeable}}}
\end{table}

Sample outputs are given in Table \ref{table:multi-voice-examples} for
the {\sc disagreeable} personality, which is one of the most distinct
in terms of aggregation and pragmatic marker insertion, along with
occurrence counts (frequency shown scaled down by 100) of the
operations that it does most frequently: specifically, {\it period
  aggregation} and {\it expletive pragmatic markers}.  Rows 1-2 shows the
counts and an example of each personality on its own. The combined personality output 
is shown in Row 3. We can see from the table that while {\sc conscientious} on its own realizes
the content in two sentences, period aggregation is much more
prevalent in the {\sc disagreeable + conscientious} example, with the
same content being realized in 5 sentences. Also, we see that some of the expletives 
originally in {\sc disagreeable} are dropped in the combined output. This suggests that the
model learns a combined representation unlike what it has seen in train, which we will explore 
in future work.

\section{Related Work and Conclusion}
\label{related-sec}

The restaurant domain has long been a testbed for conversational
agents with much earlier work on NLG
\cite{Howcroftetal13,StentPrasadWalker04,devillers2004french,gavsic2008training,MairesseBagel,Higashinakaetal07}, 
so it is not surprising that recent work using neural generation
methods has also focused on the restaurant domain
\cite{Wenetal15,Mei2015,DusekJ16,Lampouras2016,Juraska18}.  The
restaurant domain is ideal for testing generation models because
sentences can range from extremely simple to
more complex forms that exhibit discourse relations such
as justification or contrast \cite{StentPrasadWalker04}. Most recent work 
focuses on achieving semantic fidelity for simpler
syntactic structures, although there has also been
a focus on crowdsourcing or harvesting training data that exhibits
more stylistic variation \cite{novikova2017e2e,Nayaketal17,Oraby17}.

Most previous work on neural stylistic generation has been
carried out in the framework of ``style transfer'': this work is
hampered by the lack of parallel corpora, the difficulty of evaluating
content preservation (semantic fidelity), and the challenges with
measuring whether the outputs realize a particular style.
Previous experiments attempt to control the sentiment and verb tense
of generated movie review sentences \cite{hu2017toward}, the content
preservation and style transfer of news headlines and product review
sentences \cite{fuetal18}, multiple automatically extracted style
attributes along with sentiment and sentence theme for movie reviews
\cite{Ficler17}, sentiment, fluency and semantic equivalence
\cite{shen2017style}, utterance length and topic
\cite{fancontrollable}, and the personality of customer care
utterances in dialogue \cite{Herzig2017}. However, to our knowledge, 
no previous work evaluates simultaneous achievement of multiple targets as we do. 
Recent work introduces a
large parallel corpus that varies on the formality dimension, and
introduces several novel evaluation metrics, including a custom
trained model for measuring semantic fidelity
\cite{RaoTetreault18}. 

Other work has also used context representations, but not in the way
that we do here. In general, these have been used to incorporate a
representation of the prior dialogue into response generation. Sordoni
et al. \shortcite{sordoni2015neural} propose a basic approach where
they incorporate previous utterances as a bag of words model and use a
feed-forward neural network to inject a fixed sized context vector
into the LSTM cell of the encoder. Ghosh et
al. \shortcite{ghosh2016contextual} proposed a modified LSTM cell with
an additional gate that incorporates the previous context as input
during encoding. 
Our context representation encodes stylistic
parameters.

This paper evaluates the ability of different neural
architectures to faithfully render the semantic content of an
utterance while simultaneously exhibiting stylistic variations
characteristic of Big Five personalities. We created a novel parallel
training corpus of over 88,000 meaning representations in the
restaurant domain, and matched reference outputs by using an existing
statistical natural language generator, {\sc Personage}
\cite{MairesseWalker10}.  We design three neural models that
systematically increase the stylistic encodings given to the network,
and show that {\sc Model\_Context} benefits from the greatest explicit
stylistic supervision, producing outputs that both preserve semantic
fidelity and exhibit distinguishable personality styles.

\bibliography{naaclhlt2018,nl,phd}
\bibliographystyle{acl_natbib}

\end{document}